\documentclass[lettersize,journal]{IEEEtran}
\usepackage{amsmath,amsfonts,amsthm}
\usepackage{algorithmic}
\usepackage{algorithm}
\usepackage{array}
\usepackage[caption=false,font=normalsize,labelfont=sf,textfont=sf]{subfig}
\usepackage{textcomp}
\usepackage{stfloats}
\usepackage{url}
\usepackage{verbatim}
\usepackage{graphicx}
\usepackage{cite}
\usepackage[export]{adjustbox}
\usepackage{xcolor}
\usepackage[normalem]{ulem}
\usepackage{subfig}

\newtheorem{prop}{Proposition}

\begin{document}
	
	\title{Fast and Low-Memory Deep Neural Networks Using Binary Matrix Factorization}
	
	\author{Alireza Bordbar, Mohammad Hossein Kahaei ~\IEEEmembership{}
		% <-this % stops a space
		\thanks{This paper was produced by the IEEE Publication Technology Group. They are in Piscataway, NJ.}% <-this % stops a space
		\thanks{Manuscript received April 19, 2021; revised August 16, 2021.}}

	% The paper headers
	\markboth{Journal of \LaTeX\ Class Files,~Vol.~14, No.~8, August~2021}%
	{Shell \MakeLowercase{\textit{et al.}}: A Sample Article Using IEEEtran.cls for IEEE Journals}
	
	\IEEEpubid{\begin{minipage}{\textwidth}\ \\[12pt] \centering
			0000--0000/00\$00.00~\copyright~2021 IEEE
	\end{minipage}}

	\maketitle
	
	\begin{abstract}
		Despite the outstanding performance of deep neural networks in different applications,  they remain computationally expensive and require a great amount of memory. This motivates more research on reducing the resources needed for implementing such networks. An efficient approach addressed for this purpose is matrix factorization, which has been shown to be effective on different networks. In this paper, we design a training algorithm which utilizes binary matrix factorization and show its efficiency in reducing the required resources in deep neural networks. In effect, this technique leads to the fast and practical implementation of such networks.
	\end{abstract}
	
	
	\section{Introduction}
	The remarkable success of Deep Neural Networks (DNNs) has made them the go-to choice for tackling many problems in different fields. Convolutional Neural Networks (CNN) \cite{ImageNet}-\cite{ResNet} and Long Short-Term Memory (LSTM) networks \cite{LSTM1}, \cite{LSTM2} are the most common types of networks used for visual recognition and language modeling tasks. The success of DNN can be attributed to over-parametrization, which leads to computational and memory complexity. Accordingly, different methods are employed to alleviate these burdens and make state-of-the-art networks to be more practical.

	Early network compression methods are based on removing parameters with insignificant contribution to the network, commonly referred to as "pruning" \cite{Pruning1}, \cite{Pruning2}, \cite{Networksurgery}, \cite{StructuredSparsity}. While effective in reducing the number of parameters, the performance of these methods is largely dependent on the initial stages of pruning, meaning that the hyper-parameters play a great role in the success of pruning methods. Another strategy reported is based on reducing the number of parameters by applying $l_1$ regularization to the network parameters \cite{Regularization}.

	Reducing the precision of network parameters is another effective way of mitigating the computation and memory costs of deep neural networks. Training neural networks is usually done in \textit{single-precision} (FP32) arithmetic. However, \textit{half-precision} (FP16) arithmetic can also be utilized without losing much accuracy \cite{Mixed-Precision}. A combination of both FP32 and FP16 parameters can be also used to improve the performance at the cost of reducing compression. While using mixed training does not change the number of Floating-Point Operations (FLOPs) in a network, it reduces the computational cost of each operation \cite{NVIDIA}.

	Binary Neural Networks (BNN) are a special type of lower precision networks that exhibit low computational and memory costs. The main goal of these methods is to restrain the values of the network parameters to +1 and -1. This is commonly utilized in CNNs, where convolution operation accounts for a large portion of total operations. The most important challenge in these methods is the training stage, for which different approaches have been addressed. In \cite{Binarized}, a constraint is imposed on the value of gradients and then trains the network. \cite{Real-to-Binary-Convolutions} combines binary convolutions and residual training to increase the performance of the network. To compensate for the accuracy loss caused by using binary parameters, other methods have been proposed by focusing on: minimizing the quantization error \cite{XORNet}, \cite{LQNet}, modifying and improving the loss function \cite{Distilled}, \cite{Distillation and Quantization}, \cite{Apprentice}, and reducing the gradient estimation error \cite{Circulant}, \cite{Half-wave}. While these methods seem to increase the performance of the networks, there still exists a considerable gap between BNNs and real-valued CNNs.
	
	Another recent line of research focuses on exploiting the  representation redundancy of neural networks to reduce the number of parameters. As such, matrix and tensor factorization methods  have been effective in different structures, especially in CNN and Recurrent Neural Networks (RNN)\cite{Fac1}-\cite{Fac5}. One of the most recent methods  uses the multiplication of sparse matrices as a substitute for the original matrices \cite{Main}. Applying this method yields sparse networks whose accuracy is very close to that of the original, over-parametrized networks. While matrix factorization and pruning methods are distinctly different methods, both attempt to compress DNNs by the means of sparsification.
	
	In this paper, we propose the "Fast and Low-Memory Dep Neural Networks" method (FLM-DNN) by implementing binary matrix factorization (BMF) \cite{Binary1}, \cite{Binary2} in DNNs, which leads to significant reduction in computational complexity and memory requirements.

	The paper is organized as follows. In section II the proposed method is introduced and explained. In section III, the performance metrics adopted to compare the performance of different methods are introduced. Section IV is dedicated to simulations and experiments, which are conducted on a wide variety of DNNs.
	%\end{enumerate}
	
	%\begin{enumerate}
	%	\item We present a novel method of implementing DNNs with binary-factorized matrices.
	%	
	%	\item We show the effectiveness of our framework on different networks, and compare the performance of our proposed method with others.
	%\end{enumerate}

	\section{Proposed Method}
	
	\subsection{Notation}
	Similar to \cite{Main}, we use $\Theta$, $\Theta_d$, and $\Theta_c$ to show the index set of all weight matrices, the weight matrices to be factorized, and the weight matrices to be left unfactorized, respectively. The forward mapping of a network denoted by $F(\mathbf{y}|\mathbf{x})$ is referred to as a function that maps the inputs to the respective outputs, where $\mathbf{x}$ and $\mathbf{y}$ denote the input and output vectors, respectively. Also, ${\mathbf{\{W\}_i}}$ shows the set of weight matrices of the network. Then, the training process is defined by the following optimization problem:
	\begin{equation}\label{eq:training optimization}
		\begin{split}
			\underset{\mathbf{\{W_i\}}}{\text{min}} \sum_{i \in \Theta}L(p(y_n |  x_n)&,F(\mathbf{y}|\mathbf{x}; {\mathbf{\{W_i\}}}) ) \\
			+ \sum_{i \in \Theta} R(\mathbf{W}_i),
		\end{split}
	\end{equation}
	\noindent where $L(\cdot)$ denotes the loss function and \emph{R($\cdot$)} is a regularization function which can be different depending on the task and application.
	
	%Every matrix $\mathbf{W}_k \in \{\mathbf{W}_i\}$ admits binary factorization with an overwhelmingly high probability \cite{Binary2} as

	%Notice that $\mathbf{Z}_i$ includes only $\{ 0,1\} \in \{\mathbb{R}\}^{n \times r}$, where $\mathbf{A}_i$ ...(in which case the loading matrix will have a different value VAGUE PART????). It has been shown that there exists a mapping between (these two cases UNCLEAR?????), and that binary factorization is unique \cite{Binary1}, \cite{Binary2}.%
	
	%\textcolor{blue}{Matrix $\mathbf{W}_i$ can also be factorized as $\mathbf{W}_i = \mathbf{S}_i \mathbf{B}_i$, where $\mathbf{S}_i= \{\pm1\}^{n \times r}$ and $\mathbf{B}_i \in \mathbb{R}^{r \times m}$. This factorization is referred to as "sign component factorization", and there exists a unique mapping between binary and sign component factorizations. \cite{Binary1}, \cite{Binary2}. }
	
	\subsection{Training Binary-Factorized Deep Neural Networks}
	
	\begin{figure}
		\centering
		
		\subfloat[\label{fig:normal layer}]{\includegraphics[width =0.15\textwidth, valign=c]{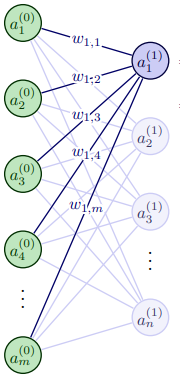} } \hfil
		\subfloat[\label{factorized layer}]{\includegraphics[width =0.25\textwidth, valign=c]{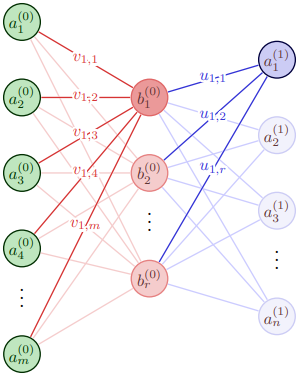} }
				
		\caption{Factorizing a fully-connected layer; (a) two layers with the original weight matrix to be factorized, (b) factorized weights by inserting $r$ neurons between the two layers. }
		\label{fig: DNN layer}
	\end{figure}
		
Matrix factorization in DNNs can be used to train a network whose weights in different layers are factorized to low-rank matrices to have simpler structures with fewer computations. Consider a weight matrix between two layers with \emph{n} and \emph{m} neurons, as shown in Fig. \ref{fig:normal layer}, for which the output vector of the second layer is given by
	
	\begin{equation} \label{eq: DNN layer}
		\begin{split}
			a_{1}^{(1)}  &= \sigma ( w_{1,1} a_{1}^{(0)} + w_{1,2} a_{2}^{(0)} + ... +w_{1,m} a_{m}^{(0)} )\\
			&= \sigma \left(\sum_{i=1}^m w_{1,i} a_{i}^{(0)}\right), \\
			\begin{pmatrix}
				a_{1}^{(1)} \\
				a_{2}^{(1)} \\
				\vdots \\
				a_{n}^{(1)}
			\end{pmatrix}
			=\sigma &\left[
			\begin{pmatrix}
				w_{1,0}  & \ldots & w_{1,m} \\
				w_{2,0}  & \ldots & w_{2,m} \\
				\vdots    & \ddots & \vdots  \\
				w_{n,0} & \ldots & w_{n,m}
			\end{pmatrix}
			\begin{pmatrix}
				a_{1}^{(0)} \\
				a_{2}^{(0)} \\
				\vdots \\
				a_{n}^{(0)}
			\end{pmatrix}
			\right],
			\\
			\mathbf{a}^{(1)}  & = \sigma \left(\mathbf{W}^{(0)} \mathbf{a}^{(0)}\right),
		\end{split}
	\end{equation}
	in which we have incorporated the bias term into the weight matrix. From now on, we replace the superscripts with subscripts to refer to the number of the layer.
	
 To factorize the weight matrix in Fig. \ref{fig:normal layer}, we insert $\emph{r}$ neurons with the identity activation function $f(x)=x$ between the two layers as shown  in Fig. \ref{factorized layer}. Thus, we get
		\begin{equation}\label{eq: factorized layer}
		\mathbf{a}_{1} =  \sigma(\mathbf{U}_0 (\mathbf{V}_0  \mathbf{a}_0 )),
	\end{equation}
		where $\mathbf{W}_0 = \mathbf{U}_0 \mathbf{V}_0$ with $\mathbf{U}_0 \in \mathbb{R}^{n\times r}$ and $\mathbf{V}_0 \in \mathbb{R}^{r \times m}$. \par

		The main goal of this work is to represent DNNs with binary-factorized weight matrices or convolution kernels, that is
			\begin{equation}\label{binaryfac}
		\mathbf{W}_k = \mathbf{Z}_k \mathbf{R}_k, \; \; \mathbf{Z}_k \in \{0, 1\}^{n\times r}, \; \; \mathbf{R}_k \in \mathbb{R}^{r\times m},
	\end{equation}
	where  $\mathbf{Z}_k$ and $\mathbf{R}_k$ are referred to as the binary and loading matrices, respectively.	We will show in Section III that the use of binary weight matrices and real-valued parameters in the DNN will significantly reduce the memory and computational costs.	 However, a major question that may arise is whether binary factorization can theoretically preserve the representational ability of the corresponding DNNs. To answer this question, we briefly  refer to the following theorems which detailed proofs are found in \cite{Binary1}, \cite{Binary2}.
		\newtheorem{mainthm}{Theorem}{\it
		\begin{mainthm}\label{thm2}
			 Matrix $\mathbf{B} \in \mathbb{R}^{n\times m }$ admits a trivial sign component decomposition with inner dimension $r=n$, as long as matrix $\mathbf{S} \in \{ \pm 1\}^{n \times n}$ is non-singular. Furthermore, a minimal binary factorization can be found where $r = rank(\mathbf{B}) $.  In the cases of minimal decomposition, sign decomposition is unique.
	\end{mainthm}}
	\begin{mainthm}\label{thm3}
		Assume matrix $\mathbf{B} \in \mathbb{R} ^{n\times m}$ admits a sign factorization as $\mathbf{B} = \mathbf{SR}, \; \textit{where} \; \mathbf{S} \in \{ \pm 1 \}^{n \times r} \; \text{and} \; \mathbf{R} \in \mathbb{R} ^ {r \times m} $. Then, matrix $\mathbf{C} = \frac{1}{2} \left(\mathbf{B} + \mathbf{E}\right) $ admits a binary decomposition as $\mathbf{C} = \mathbf{Z} \mathbf{R} \; \text{where} \; \mathbf{Z} \in \{0,1\}^{n\times r}$ and $\mathbf{E}$ is a matrix of ones with appropriate dimensions. Not only  loading matrices for both matrices $\mathbf{B}$ and $\mathbf{C}$ are equal, but also there exists the mapping
		$$F: \{0,1\} ^{n \times r} \rightarrow \{\pm 1 \}^{n \times r} ;
		F: \mathbf{Z} \rightarrow 2\mathbf{Z} - \mathbf{E}$$
		between the sign and binary matrices $\mathbf{S}$ and $\mathbf{Z}$.
\end{mainthm}
		Thus, to represent a DNN with factorized weights matrices, according to (\ref{binaryfac}), (\ref{eq:training optimization})   is modified as
	\begin{equation} \label{Bayesian}
		\begin{split}
			\{\mathbf{W}_i^\star\} = & \underset{\{\mathbf{W}_i\} }{
				\text{arg max}} \: \mathbb{E}_{\mathbf{x} , \mathbf{y}} \left[ \text{log} \left(p \left( \left\{\mathbf{W}\right\}_{i} | \mathbf{x}, \mathbf{y}
			\right)
			\right)
			\right] \; , \\
			&\text{subject to} \; \mathbf{W}_i = \mathbf{Z}_i \mathbf{R}_i \;  ,\mathbf{Z}_i \in \{0,1\}^{n\times r}, \; \; \\
			&\mathbf{R}_i \in \mathbb{R}^{r\times m}.
		\end{split}
	\end{equation}
	It is immediately obvious that training a network using (\ref{Bayesian}) is very challenging due to the fact that the imposed binary constraint on matrices $\mathbf{Z}_i$ is not convex. Instead, an equivalent convex problem can be defined as follows, which enables us to train a network with sign-factorized weights first, and then map the sign matrices to binary matrices:
		\begin{equation}\label{equivalent}
		\begin{split}
			\{\mathbf{W}_i^\star\} = & \underset{\{\mathbf{W}_i\} }{
				\text{arg max}} \: \mathbb{E}_{\mathbf{x} , \mathbf{y}} \left[ \text{log} \left(p \left( \left\{\mathbf{W}\right\}_{i} | \mathbf{x}, \mathbf{y}
			\right)
			\right)
			\right] \; , \\
			&\text{subject to} \; \mathbf{W}_i = \mathbf{S}_i \mathbf{A}_i\; \\& -1 \leq s_{jk} \leq 1, \; \; \\ &\|\mathbf{S}\|_F^2 = n \times r \; \; for \; \forall\: \mathbf{S} \in \left\{\mathbf{S}\right\}_i.
		\end{split}
	\end{equation}

	According to (\ref{equivalent}), as long as weight matrix entries remain within $\left[-1,1\right]$, and the Frobenius norm of each sign weight matrix in $\left\{\mathbf{S}_i\right\}$ is equal to the square root of the number of its entries, after training the sign-factorized weights will be obtained. 	However, although sign matrix factorization yields a network with a significantly lower memory consumption than real-valued networks, it is ineffective in reducing the number of FLOPs compared to sparse DNN compression methods such as \cite{Main}. Subsequently, after the sign-factorized matrices are determined, we map the sign matrices of each factorized matrix to binary matrices according to Theorem \ref{thm3}. In this way, in the \emph{k}-th layer of the network, the relationship between the input $\mathbf{x}_k$ and output $\mathbf{y}_k$ is given by
		\begin{equation} \label{eq:sign to binary in layer}
			\begin{split}
				\mathbf{y}_k &= \sigma \left(\mathbf{S}_k \mathbf{R}_k \mathbf{x}_k\right) = \sigma \left(\mathbf{B}_k \mathbf{x}_k \right) = \sigma \left[ \left(2\mathbf{C}_k - \mathbf{E}\right) \mathbf{x}_k\right]  \\
			&	= \sigma \left[2 \mathbf{Z}_k \mathbf{R}_k \mathbf{x}_k -\mathbf{E}\mathbf{x}_k\right].
			\end{split}			
		\end{equation}
In other words,	according to (\ref{eq:sign to binary in layer}), a sign-factorized representation is converted to binary-factorized representation, where the term $-\mathbf{E} \mathbf{x}_k$ is added as a bias vector to the neurons of the next layer. The FLM-DNN method is shown in Algorithm \ref{alg:FLM-DNN}.
			\begin{algorithm}[H]
		\caption{Neural Network with Binary Factorization Training. }\label{alg:FLM-DNN}
		Initialize the weight matrices $\mathbf{Z}_i$ and $\mathbf{R}_i \in \Theta_d$; initialize learning rate $\lambda_0$; initialize $\mathbf{W}_i \in \Theta_c$; set the number of training epochs T; collect the set of training sample $\left\{\left(x_{n'}, y_{n'}\right)\right\}$.
		\begin{algorithmic}[1]
			\FOR{t=1,..., T}
			\STATE $\lambda$ $\gets$ $\lambda_0$ $\times$ $g((t-T_0)/T_1)$
			\REPEAT
			\STATE Collect a mini-batch of samples from $\left\{\left(x_{n'}, y_{n'}\right)\right\}$.
			\REPEAT
			\STATE If $i \in \Theta_c$, update $\mathbf{W}_i$ using back propagation. Else, update $\mathbf{Z}_i$ with norm projection and $\mathbf{R}_i$ normally.
			\UNTIL{All learnable parameters are updated.}
			\UNTIL{All mini-bathches are used to train}
			\ENDFOR
			\STATE $\mathbf{Z}_i$ $\gets$ $(\mathbf{Z}_i+1)/2, \; \forall i \in \Theta_d$ 
			\STATE $\mathbf{R}$ $\gets$ $2\mathbf{R}, \; \forall i \in \Theta_d $.
			\STATE Add $-\mathbf{E}_{i-1} \mathbf{x}_{i-1}$ to the \textit{i}-th layer neurons as bias $\forall i \in \Theta_d$.
			\STATE output $\left\{\mathbf{W}_i\right\}$, $\left\{\mathbf{R}_i\right\}$, $\left\{\mathbf{Z}_i\right\}$
		\end{algorithmic}
		\label{alg1}
	\end{algorithm}
	
	 %\sout{To ensure a smooth and stable training, the following procedures are  incorporated into the training algorithm:} \textcolor{blue}{THE ITEMS WILL BE REMOVED!}
	%[\begin{itemize}
	%	\item The initialization of binary weight matrices is done by randomly selecting the entries from $\left\{-1 , +1\right\}$ with  the probability of $p=0.5$ for the selection of $+1$. This ensures that the average value of the entries is close to zero.
		%	\item The regularization term for binary components is modified to
			%	$$ R(\mathbf{x})= \alpha ||\mathbf{x}+1||_1,$$
			%	which drives the weights towards -1.
	%	\item Frobenius norm projection is done for each mini-batch of training data.
	%	\end{itemize}]

	%\sout{Finally, a point to be discussed is applying a threshold function addressed in} \cite{Main} \sout{to turn the real-valued weight matrices into ternary-valued matrices, which causes a noticeable accuracy drop. However, we noticed in our experiments that this is not the case with our proposed method, as the performance of the networks is surprisingly preserved well.}
	
	Another point to be emphasized is that although \cite{Main} proposes using a threshold function to turn the real-valued weight matrices into ternary-valued matrices, the resulting accuracy drop is significant compared to the real-valued case. We will show in our experimental results that this is not the case with our method, as the network surprisingly performs well in different scenarios.

	\section{Memory Requirements and Computational Complexity}
	One convenient metric to compare the complexity of different networks and algorithms is the compression rate, which is defined as the ratio of total parameters in the original network to non-zero parameters in the compressed network, that is,
	\begin{equation} \label{sparsity}
		\textit{Compression rate} = \frac{\textit{Total number of parameters}}{\textit{Number of non-zero parameters}}.
	\end{equation}
	While compression rate is a sensible metric when comparing the performance of real-valued networks, it is incapable of taking into account the effectiveness of binary and real weights in memory and computation reduction. This will be highlighted more in our work due to the nature of binary factorization, where the binary component can be stored using 1 bit rather than the traditionally used 32 bits, leading to lower memory consumption and faster computations. As a result, for comparison purposes,  we use a more relevant metric presented in  \cite{Bi-Real}, \cite{ResNet} to evaluate the number of memories required to store the network and the number of FLOPs in the forward pass.
	%In this paper, however, we use more relevant metrics to compare memory and computation efficiency of different networks and algorithms. Due to the nature of binary factorization, the sign component can be stored using 2 bits rather than the more traditional 32 bits, which leads to lower memory consumption and faster computation. While (\ref{sparsity}) is a sensible metric when comparing the performance of real-valued networks, it is incapable of taking into account the effectiveness of binary and real weights in memory and computation reduction. Therefore, similar to \cite{Bi-Real}, \cite{ResNet}, we compare the memory required to store the network, and the number of floating-point operations (FLOPs).
	%Calculating memory is straightforward: real-valued parameters each occupy 32 bits of memory, while binary weights only occupy one.

	\begin{equation}
		\begin{split}
			\textit{Total Memory} = & \textit{ (number of real-valued params.) } \times 32\\
			& + \textit{the number of binary params.}.\\
		\end{split}
	\end{equation}
	
	Also, to evaluate the computational complexity  of different networks experimented in this work, the number of FLOPs is calculated using the following guidelines.

	For  dense, real-valued networks, the number of FLOPs can be calculated  as follows:
	
	\begin{itemize}
		\item In a fully-connected layer, it equals $2 \times \textit{(the number of neurons in current layer)} \times \textit{(the number of neurons in previous layer)}$.
		
		\item In a convolutional layer, it equals $2 \times \textit{(the number of input kernels)} \times \textit{(kernel shape)} \times \textit{(the number of kernels in current layer)} \times \textit{(output shape)}$.
	\end{itemize}
	In recurrent networks such as vanilla RNN (Recurrent Neural Netork), GRU (Gated Recurrent Unit), and LSTM (Long Short-Term Memory) wherein the dominant operation is matrix-vector factorization and vector addition, the number of FLOPs is calculated in a similar way to multi-layer perceptron networks.
	
	When matrix factorization is applied to a layer, the number of FLOPs is obtained as follows \cite{Main}:
	
	\begin{itemize}
		\item When a weight matrix $\mathbf{W} \in \mathbb{R}^{n \times m}$ between two fully-connected layers is factorized into two dense and real-valued matrices $\mathbf{A} \in \mathbb{R}^{n \times r}$ and $\mathbf{B} \in \mathbb{R}^{r \times m}$, it is modified to
		$$\text{FLOPs}=   2\times \left(\frac{nr}{t_A} + \frac{mr}{t_B} \right),$$
		where $t_A$ and $t_B$ are the compression rates of $\mathbf{A}$ and $\mathbf{B}$, respectively. When the latter matrices are not sparse, their compression rates are set to 1.
		
		\item When a convolutional kernel is factorized into two kernels, it is calculated by dividing the number of FLOPs in new kernels by their respective compression rates. Convolutional layer factorization is performed according to \cite{ConvFac}.
	\end{itemize}
	
	Lastly, when binary representation is used along with sparse factorization, the number of additions and multiplications in the network should be calculated separately and added.
	
	The number of parameters in the networks used to experiment on here can be calculated using the following formulae.
	\begin{itemize}
		\item In a dense  fully-connected layer, the number of parameters equals $\textit{(the number of neurons in the current layer)} \times \textit{(the number of neurons in the previous layers)}.$
		\item In a dense convolutional layer,  the number of parameters is
		\begin{equation*}
			((m \times n) \times d) + 1) \times k
		\end{equation*}
		where  $k$ and $d$ are is the number of filters in a layer and its previous layer, respectively, and $m \times n$ is the shape of the kernels in the current layer.
		\item A vanilla RNN having three layers with \emph{m} input, \emph{n} output, and \emph{h} hidden units, the number of parameters, including two bias vectors, at each time stamp is equal to
		\begin{equation*}
			h^2 + nh + mh + n + h.
		\end{equation*}
		Similarly, a GRU network with \emph{m} input and \emph{n} output units has $3(n^2 + nm + 2n)$ parameters. Finally, in an LSTM network with the same number of inputs and outputs, the total number of parameters is $4 (n^2 + nm +n)$.
	\end{itemize}

	\section{Experiments}
	In this section, the effectiveness of FLM-DNN algorithm is evaluated on a wide variety of networks and some commonly used datasets, and its performance is compared with other state-of-the-art network compression methods.  %And probably tensorflow and one other%

	\subsection{Setup and Datasets}
	We experimented on nine different networks including LeNet-300-100, LeNet-5 Caffe \cite{Caffe}, a VGG-like network\footnote{htttps://github.com/geifmany/cifar-vgg.git}, ResNet164, ResNet50, Vanilla RNN, two LSTM networks, and one GRU network.
	The datasets are differently selected according to the task and network. Overall, we use five datasets in the experiments including: MNIST \cite{MNIST}, CIFAR-10 \cite{CIFAR}, CIFAR-100 \cite{CIFAR}, ImageNet \cite{ImageNet}, and Penntreebank (PTB) \cite{PTB}. To introduce, the MNIST  consists of 70000  labeled gray images,  CIFAR-10 and CIFAR-100 consist of 60000 color images with 10 and 100 classes, respectively, ImageNet is a annotated color image data set of  14,197,122 images. Also, the Penn Treebank (PTB) project selected 2,499 stories from a three year Wall Street Journal (WSJ) collection of 98,732 stories for syntactic annotation, which is widely used for Natural Language Processing (NLP) applications. All the data sets are split according to the respective original references.
	
	For Multi-Layer Perceptron (MLP) networks, LeNet-300-100 is used, which is a four-layer fully-connected network with two hidden layers containing 300 and 100 neurons, respectively. The output layer consists of 10 neurons. Also, the MNIST is applied for training and evaluating the network.

	The values and settings provided here will remain the same throughout the rest of the experiments, unless stated otherwise. Adam \cite{Adam} is used as the optimizer, and the number of epochs for MLP training is set to 300. Since norm regularization methods only make the weights approach zero, a threshold value $\epsilon$ is adopted. This value is $exp(-4)$ for percepron layers, $exp(-6)$ for convolutional layers, and $exp(-4)$ for recurrent units.
	
	(Similar to \cite{Bi-Real},) We adopt memory consumption and the number of FLOPs needed in a single forward pass stage as metrics to evaluate and compare the performance of our proposed method with a few other state-of-the-art DNN compression methods.
	
	\subsection{Validation of Training Algorithm}
	In this section, we design an experiment to empirically evaluate the effectiveness of the FLM-DNN training method  to approximate a given matrix $\mathbf{W} \in \mathbb{R}^{n \times m}$ with rank \emph{r} in an MLP network with only one factorized layer.
To do so, consider a neural network with \textit{m} input and \textit{n} output neurons and one hidden layer. Furthermore, assume the hidden and output layers have identity activation function $\sigma(X)=x$. Then, using binary factorization according to  (\ref{eq: factorized layer}) and (\ref{binaryfac}), the relationship between the input vector $\mathbf{x} \in \mathbb{R}^m$ and output vector $\hat{\mathbf{y}} \in \mathbb{R}^n$ can be written as
	\begin{equation}
		\hat{\mathbf{y}} = \mathbf{Z_*}\mathbf{R_*} \mathbf{x}
	\end{equation}
	where $\mathbf{Z_*} \in \{0,1\}^{n \times r} $ and $\mathbf{R_*} \in \mathbb{R}^{r \times m}$.
	Also, assume that we utilize this network for a linear regression problem. Therefore, the loss function over a mini-batch $\{\mathbf{x}_{n'} , \mathbf{y}_{n'}\}$  of the training data with $N'$ data points is defined as
	\begin{equation}\label{eq: loss function}
		\begin{split}
			L\left( \hat{\mathbf{y}}, \mathbf{y}  \right) & = \frac{1}{N'}  \sum_{n'=0}^{N'-1} \|\hat{\mathbf{y}}_{n'} - \mathbf{y}_{n'}\|_2^2 \\
			&  = \frac{1}{N'}  \sum_{n'=0}^{N'-1} \| \mathbf{Z_*}\mathbf{R_*} \mathbf{x}_{n'} - \mathbf{y}_{n'}\|_2^2.
		\end{split}
	\end{equation}
	Then, we can design the data set in a way that the loss function in  (\ref{eq: loss function}) achieves its minimum value for  $\mathbf{Z_*R_*} = \mathbf{W}$.
	
	\begin{prop}
For a set of linearly independent vectors $\{\mathbf{x}_{n'}\}$, by setting $\mathbf{y}_{n'} = \mathbf{W}\mathbf{x}_{n'}$, the loss function in (\ref{eq: loss function}) achieves its minimum for $\mathbf{Z_*R_*}=\mathbf{W}$.
	\end{prop}
	\begin{proof}
		Computing the gradient of $L(\hat{\mathbf{y}} , \mathbf{y})$ with respect to $\mathbf{y}$ yields
		\begin{equation}\label{eq: gradient}
		\nabla_{\mathbf{y}}  L(\hat{\mathbf{y}} , \mathbf{y}) = \frac{-2}{N'} \sum_{n'=0}^{N'-1} \left(\mathbf{Z_*}\mathbf{R_*} \mathbf{x}_{n'}  - \mathbf{y}_{n'} \right),
		\end{equation}
		and setting $\mathbf{y}_{n'} = \mathbf{W}\mathbf{x}_{n'}$ reduces $\nabla_{\mathbf{y}}  L(\hat{\mathbf{y}} , \mathbf{y})$ to zero. As  $L(\hat{\mathbf{y}} , \mathbf{y})$ is a convex function in $\mathbf{y}$, this solution achieves the global minimum. This concludes the proof.
	\end{proof}

We conduct Monte-Carlo simulations to verify Proposition 1 and the representational capability of FLM-DNN. In each iteration,  $\mathbf{Z} \in \{0,1\}^{n \times m}$ and $\mathbf{R} \in \mathbb{R}^{r \times m}$ are generated randomly using Bernoulli and Gaussian  distributions, respectively. Next, matrix $\mathbf{W} = \mathbf{ZR}$ is constructed. Then, $2^{18}$ input vectors $\{\mathbf{x}_i\}$ are randomly drawn from a Gaussian distribution, and the output vectors $\{\mathbf{y}_i\} = \{\mathbf{Wx}_i\}$ are calculated. Finally, the corresponding input and output vectors are paired together as $\{(\mathbf{x}_i , \mathbf{y}_i)\}$ and the network is trained using Algorithm \ref{alg:FLM-DNN}. This process is repeated 20 times, and the mean results are reported in Tables \ref{table: verification fixed rank} and \ref{table: verification fixed dimensions}.
First, we investigate the construction error of matrices with fixed rank and varying sizes. The results are shown in Table \ref{table: verification fixed rank}, in which the reconstruction error (RE) is defined as
\begin{equation} \label{eq : reconstruction error}
	RE = \frac{\| \mathbf{W} - \mathbf{W}_*\|_F}{\| \mathbf{W} \|_F}
\end{equation}
for reconstructed matrix $\mathbf{W}_*$.
 It is observed that for low-rank matrices, the FLM-DNN training algorithm approximates $\mathbf{W}$ very accurately, in the sense that a reconstruction error of 0.01 causes almost indistinguishable differences visually, as will be demonstrated in Section IV-C.

\begin{table}
	\centering
	\caption{Reconstruction error of matrix $\mathbf{W}$ by FLM-DNN for a fixed rank  and varying dimensions.}
	\begin{tabular}[!t]{c c c}
		Dimensions& Rank & RE \\
		\hline
	
		$50 \times 25 $ & 10 & 8e-3 \\
		$100 \times 50$ &10 &7e-5\\
		$150 \times 75$ & 10& 8e-5\\
		$200 \times 100$ &10& 1e-4\\
		$300 \times 150$ & 10& 2e-4\\
		\hline
	\end{tabular}
\label{table: verification fixed rank}
\end{table}

\begin{figure}[!h]
	\centering
	\includegraphics[width=3.3in]{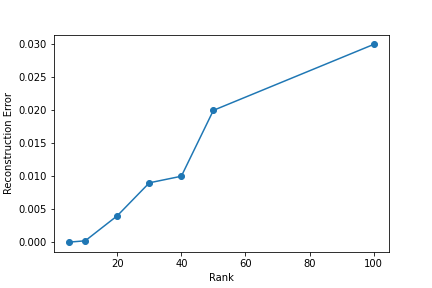}\\
	\caption{Plotted reconstruction error (
		$\frac{\| \mathbf{W} - \mathbf{W}_* \|}{\| \mathbf{W} \|}$
		) of matrices with dimensions $300 \times 150$ and varied ranks from table \ref{table: verification fixed dimensions}. The reconstruction error increases with rank.}
	\label{fig:reconstruction error}
\end{figure}

The second experiment focuses on the case where the dimensions of matrix $\mathbf{W}$ are fixed to $n=300$ and $m=150$, but its rank varies. The results of this experiment, as demonstrated in Table \ref{table: verification fixed dimensions} and Fig. \ref{fig:reconstruction error}
  suggests that as the rank of $\mathbf{W}$ increases, the reconstruction error increases as well. This is due to the combinatorial and discrete nature of binary matrix factorization, meaning that as the matrix rank increases,  the number of independent binary vectors that should be correctly found by the network increases as well. In other words, this may lead to an increasing number of incorrectly identified binary vectors, which leads to increasing reconstruction errors.

\begin{table}
	\centering
	\caption{Reconstruction error of matrix $\mathbf{W}$ by FLM-DNN for a fixed dimension and varying ranks.}
	\begin{tabular}[!h]{c c c}
		 Dimension & Rank & RE \\
		\hline
		
		$300 \times 150$ &5& 1.26e-6 \\
		$300 \times 150$ &10 & 2e-4\\
		$300 \times 150$ & 20 & 4e-3\\
		$300 \times 150$&30 & 9e-3\\
		$300 \times 150$& 40 & 1e-2\\
		$300 \times 150$&50 & 2e-2\\
			$300 \times 150$&100 & 2e-2\\
		\hline
	\end{tabular}
\label{table: verification fixed dimensions}
\end{table}

\subsection{Image Reconstruction with Autoencoders}
In the first practical experiment, we employ the FLM-DNN in image reconstruction using an autoencoder network, which   consists of 7 hidden layers with 256, 128, 64, 10, 64, 128, and 256 neurons. The dataset used is MNIST, and the output layer has 10 neurons and utilizes softmax activation.
\begin{figure*}
	\centering
	\begin{tabular}{c c}
		\rotatebox[origin=c]{0}{ \small FLM-DNN} &
		\includegraphics[width=16cm , valign=c]{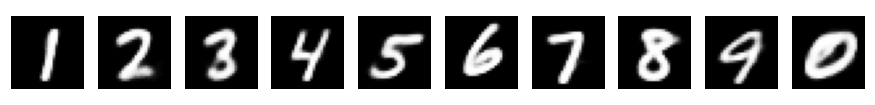} \\
		\rotatebox[origin=c]{0}{\small SMF \cite{Main}} &
		\includegraphics[width=16cm, valign=c]{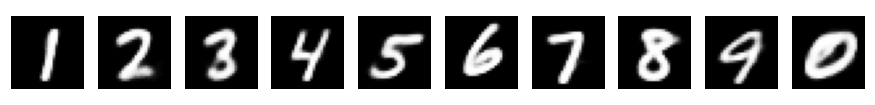} \\
		\rotatebox[origin=c]{0}{\parbox{1cm}{\small \centering Original  Images \cite{MNIST}}} &
		\includegraphics[width=16cm, valign=c]{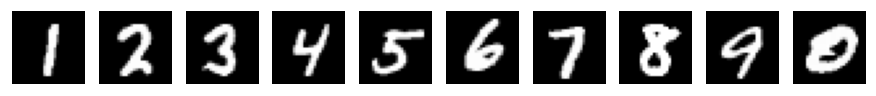}
	\end{tabular}
	\caption{Comparison between the performance of SMF \cite{Main} and FLM-DNN in image reconstruction. Visually, the difference between SMF and FLM-DNN is not noticeable.}
	\label{reconstruction}
\end{figure*}

 The two weight matrices with dimensions $ 764 \times 256$ are factorized with a typical dim        ension of p = 250. $L_1$  norm regularization is used in all layers to sparsify the network. After training, 10 images with 10 different labels are chosen from the test data and plotted. Fig. \ref{reconstruction} demonstrates the constructed images and Table \ref{table:reconstruction} compares the memory consumption, computational complexities, and accuracies with the original
dense network and the SMF algorithm \cite{Main}. As seen in Fig. \ref{reconstruction} and Table \ref{table:reconstruction}, the reconstruction error rates for SMF and FLM-DNN are very close, leading to almost indistinguishable differences in the reconstructed images. But the main point that highlights the merit of FLM-DNN is that it achieves these results using less memory consumption and fewer computations.
\begin{table} [!h]
	\caption{Comparison of FLOPs, memory usage, and reconstruction error an autoencoder network after being trained by SMF, FLM-DNN, and
		the original dense network}
	\centering
	\begin{tabular}{c c c c }
		\hline
		Method & Memory (MBits) & FLOPs & RE \\
		\hline
		Original network  & 16.01 &9.69e5& 0.227 \\
		SMF \cite{Main} &0.23 & 1.09e4 &0.233\\
		FLM-DNN &0.17 & 9.46e3 &0.241\\
		\hline
		
	\end{tabular}
	\label{table:reconstruction}
\end{table}
\subsection{Experiments on Multi-Layer Perceptron Networks}
To evaluate the proposed FLM-DNN in MLP networks, we investigate the accuracy and computational complexity of different algorithms in the LeNet-300-100  network \cite{MNIST}. The results are compared with the other state-of-the-art methods, such as the progressive pruning \cite{Pruning1}, Dynamic Network Surgery (DNS) \cite{Fac4}, and Sparse Variational Dropout (SVD) \cite{Fac6}. The layer-wise scaling regularization factors are selected as $\left[0.00001, 0.000025, 0.00015\right]$. Similar to \cite{Main}, a notation like "1-0-0" means that the first weight matrix is factorized, and the others are not. Table \ref{table:other methods} shows the results. 

Our next experiment investigates  the effect of factorizing different layers on accuracy and computational complexity in LeNet-300-100 network. The results are compared in Table \ref{table:config}.

Next, we investigate the effect of changing the common dimension of LeNet-300-100  network on the accuracy and computational complexity of algorithms. In this experiment, only the first layer of the network is factorized. The results are demonstrated in Table \ref{table:common dimension}.
\\
The following conclusions can be drawn from the results shown in Tables \ref{table:other methods}, V, and \ref{table:common dimension}(DO YOU MEAN THESE TABLES???). (WE SHOULD TALK???)
\begin{enumerate}
	\item FLM-DNN competes with the other state-of-the-art network compression methods in terms of accuracy, but with a lower computational cost.
	\item  As demonstrated in Table \ref{table:config}, configurations in which the first layer of the LeNet-300-100 network is factorized are more compressed than others. Furthermore, factorizing the weight matrix of the second layer leads to more compression than the third layer. Since the layers of the LeNet-300-100 network decrease in size as we move toward the output layer, this finding suggests that factorizing larger layers in an MLP network lead to more memory and computation reduction.
	\item  Reducing the rank of the factorized weight matrix will not deteriorate the performance of the network, unless the dimension is chosen very small. Furthermore, reducing the dimension does not necessarily lead to more compression, since it reduces the number of parameters in the weight matrix of a layer as well as the rank of that matrix, meaning that the achievable compression rate for that weight matrix is also reduced.
\end{enumerate}

%The regularization factors for different layers are $\left[0.00001, 0.000025, 0.00015\right]$, and "1-0-0" configuration is chosen.
\begin{table} [!h]
	\caption{Comparison of different methods in terms of memory usage, FLOPs, and error rate (ER). }
	\centering
	\begin{tabular}{c c c c }
		\hline
		Method & Memory (KBits) & FLOPs & ER (\%)\\
		\hline
		Original \cite{MNIST} &8518 & 5.32e5 &1.64 \\
		Progressive\cite{Pruning1} &696.83 & 4.36e4 &1.59 \\
		DNS \cite{Fac4} & 151.63 &9.48e3& 1.99 \\
		SVD  \cite{Fac6}& 127.51 &7.97e3& 1.92\\
		SMF $(\text{1-0-0-}l_1) $ \cite{Main}& 122.14 &7.63e3& 1.83 \\
				FLM-DNN $(\text{1-0-0-}l_1) $  & 86.52 &7.1e3& 1.85 \\
		\hline
	\end{tabular}
	\label{table:other methods}
\end{table}

\begin{table} [!h]
	\caption{Comparison of  FLM-DNN with SMF in different structures of LeNet-300-100 in terms of memory usage, FLOPs, and accuracy.}
	\centering
	\begin{tabular}{c c c c c }
		\hline
		Method &Config. & Memory (KBits) & FLOPs & ER (\%) \\
		\hline
		
		&1-1-1 &123.88 &7.69e3 & 1.70\\
		&1-1-0 &121.48 &7.58.e3 & 1.85\\
		&1-0-1 &112.83 &7.06e3& 1.83 \\
		SMF \cite{Main}	&0-1-1&151.56 & 9.47e3 &  1.80\\
		&0-1-0 & 126.91& 7.93e3 & 1.89\\\
		&0-0-1 & 135.58 & 8.48e3 &1.88\\
		&1-0-0 & 100.48 & 6.28e3 & 1.68 \\
		
		\hline
		
		&1-1-1 &48.40 &5.17e3 & 1.72\\
		&1-1-0 &50.95 &5.28e3 & 1.86\\
		&1-0-1 &64.79 &5.48e3& 1.83 \\
		FLM-DNN	&0-1-1 & 90.90 & 7.51e3 & 1.87\\
		&0-1-0 & 101.66& 7.18e3 & 1.85 \\
		&0-0-1 & 103.03 & 7.42e3 &1.86\\
		&1-0-0 & 84.88 &5.71e3 & 1.73 \\
		\hline
	\end{tabular}
	\label{table:config}
\end{table}
\begin{table} [!h]
	\caption{Comparison of  FLM-DNN with SMF for different common dimensions in LeNet-300-100 network when only the first layer is factorized.}
	\centering
	\begin{tabular}{c c c c c }
		\hline
		Method & Dimension & Memory (KBits) & FLOPs & ER (\%) \\
		\hline
		
		&50 &115.60 &7.23e3 & 2.24\\
		&100 &104.38 &6.53.e3 & 2.03\\
		&150 &106.72 &6.67e3& 2.10 \\
		SMF \cite{Main}	&200&104.28 & 6.51e3 &  1.83\\
		&217 & 106.42& 6.65e3 & 2.15\\
		&250 & 102.84 & 6.42e3 &1.99\\
		&300 & 105.39 & 6.64e3 & 2.20 \\
		
		\hline
		
		&50 &70.21 &5.71e3 & 2.27\\
		&100 &69.86 &5.34.e3 & 2.06\\
		&150 &72.28 &5.49e3& 2.08 \\
		FLM-DNN	&200& 71.44 & 5.39e3 &  1.81\\
		&217 & 71.23&5.46e3 & 2.18 \\
		&250 & 73.96 & 5.49e3 &2.03\\
		&300 & 79.22 & 5.74e3 & 2.22 \\
		\hline
	\end{tabular}
	\label{table:common dimension}
\end{table}
\subsection{Experiments on CNNs}
We have considered four CNN networks for our experiments. The first one is LeNet-5 Caffe\cite{Caffe}, one of the simplest CNNs consisting of an input layer, four convolutional layers, two pooling layers, and two fully-connected layers with 500 and 10 neurons, respectively. The training and evaluation dataset is MNIST. The next network is a simplified version of the VGG network \cite{VGG}, which shall be referred to as "VGG-Like" \cite{VGG-Like}, for which we utilize the  CIFAR-10 dataset. Furthermore, the two residual networks ResNet164 and ResNet50 \cite{ResNet} are tested and  ImageNet and CIFAR-100 datasets are used for training, respectively. Since most of the LeNet-5 parameters lie in the fully-connected layers, only those layers are factorized.  The common dimension of factorization is $p=400$, and $l_{1}$ regularization is utilized.
\begin{table}[!b]
	\caption{Performance comparison of algorithms in LeNet-5 Caffe trained with MNIST dataset.}
	\centering
	\begin{tabular}{c c c c c}
		\hline
		Network & Method & Memory (Mb) & FLOPs & ER (\%) \\
		\hline
		LeNet-5 Caffe&Original  \cite{Caffe} &13.21 &4.58e6 & 0.8 \\
		& SMF \cite{Main} & 0.048 & 3.0e5 & 0.88\\
		& FLM-DNN & 0.040 & 2.99e5 & 0.90\\
		\hline
	\end{tabular}
	\label{table:LeNet-5}
\end{table}

\begin{figure}[!h]
	\centering
	\subfloat[\label{conv1}]{\includegraphics[width =3.2in]{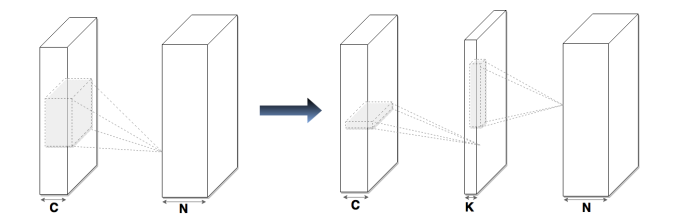}}\\
	\caption{Factorizing a convolutional layer into two layers with a lower rank. The original kernel is divided into horizontal and vertical layers\cite{ConvFac}.}
	\label{fig:convolutional factorization}
\end{figure}

From Table \ref{table:LeNet-5}, one can see that while FLM-DNN reduces the memory consumption about 17\%, the number of FLOPs almost stays the same. The reason is that, despite the fact that a huge portion of the parameters lie in the first fully-connected layer, it is the second deep convolutional layer that takes up the biggest part of the calculations. Fortunately, convolutional layers can be factorized in a similar manner to Fig. \ref{fig: DNN layer} by using horizontal and vertical filters to approximate the original filter \cite{ConvFac} as shown in Fig. \ref{fig:convolutional factorization}. We apply  FLM-DNN algorithm to the method proposed in \cite{ConvFac} to compress the convolutional layers in  VGG-like and two residual networks.

Given a convolutional layer that includes $n \times c$ filters with a common spatial size of $3 \times 3$, one can factorize them into two low-rank groups of $n \times p$ and $p\times c$ layers with the spatial sizes of $3 \times 1$ and $1 \times 3$, forming horizontal and vertical filters, respectively. In the VGG-like network, the last three convolutional layers ($n=c=512$) along with the first fully-connected layer ($m=n=512$) are factorized. The learning rate for training this network is $0.0005$. The norm regularization factor is set to  $\lambda_0 = 0.0006$ . Table \ref{table:VGG-like} demonstrates the effectiveness of FLM-DNN in a VGG-like network. As observed, by applying binary convolutional factorization, both memory consumption and the number of calculations are considerably reduced. The reason is that by binary-factorizing the last convolutional layer, the number of FLOPs are significantly reduced, which was not the case in LeNet-5 network.

\begin{table}[!h]
	\caption{Performance comparison of algorithms in the VGG-like network trained with CIFAR-10.}
	\centering
	\begin{tabular}{c c c c c}
		\hline
		Network & Method & Memory (Mb) & FLOPs & ER (\%) \\
		\hline
		VGG-like & Original \cite{VGG-Like} & 765.19 & 3.69e8 & 6.90 \\
		& SMF \cite{Main}& 104.83 & 8.7e6 & 0.88\\
		& FLM-DNN & 85.91 & 7.05e6 & 0.90\\
		\hline
	\end{tabular}
	\label{table:VGG-like}
\end{table}

The same convolutional layer factorization method is applied to ResNet-50 and ResNet164 residual networks, which consist of residual "blocks" that are repeated consecutively to form larger blocks, or "modules". More information can be found in \cite{ResNet} on how different structures of residual networks are formed.
 Here, we only factorize the convolution layers in the last module of ResNet-50 and ResNet-164. The results are shown in Table \ref{table:ResNet}. Here, similar to the VGG-Like network, factorizing the layers using FLM-DNN method has led to significant reduction in FLOPs and memory.

\begin{table}[!h]
	\caption{Performance comparison of algorithms in ResNet-50 and Resnet-164 networks trained with ImageNet and CIFAR-100, respectively.}
	%Performance Comparison of ResNet-50 and Resnet-164 Networks Trained Using ImageNet and cifar-100 Datasets, respectively, using Sparse Matrix Factorization and Binary Matrix Factorization Algorithms }
\centering
\begin{tabular}{c c c c c}
	\hline
	Network & Method & Memory  (Mb) & FLOPs & Top-1 (\%) \\
	\hline
	ResNet50   &  Original\cite{ResNet} & 733.03 & 3.8e9 & 23.51 \\
	& SMF \cite{Main} & 109.73 & 6.5e7 & 24.65\\
	& FLM-DNN & 88.88 & 5.54e7 & 24.82\\
	\hline
	ResNet164   & Original \cite{ResNet} & 54.08 & 2.6e8 & 23.71 \\
	& SMF \cite{Main} & 13.93 & 7.6e7 &  23.91\\
	& FLM-DNN &11.74 & 6.53e7 & 23.99\\
	\hline
\end{tabular}
\label{table:ResNet}
\end{table}
\subsection{Experiments on RNN, GRU, and LSTM Networks}
First, a vanilla RNN \cite{RNN} is trained using MNIST dataset. The rows of images are fed into the network sequentially. The vanilla RNN has three parametrized input-to-hidden, hidden-to-hidden, and hidden-to-output layers. However, we  only factorize input-to-hidden and hidden-to-output matrices. The GRU network is composed of seven layers including three input-to-hidden, three "hidden-to-hidden", and one "hidden-to-output" layer. All matrices except the hidden-to-output layer are factorized, $l_1$-norm is utilized, and the regularization factor is set to $0.0003$. The networks are trained for 300 epochs.
\begin{table}[!h]
\caption{Performance comparison of algorithms in Vanilla RNN trained with MNIST.}
\centering
\begin{tabular}{c c c c c}
	\hline
	Network & Method & Memory (Mb) & FLOPs & ER (\%) \\
	\hline
	Vanilla RNN &  Original \cite{RNN} & 2.65 & 4.62e6 & 1.67 \\
	& SMF \cite{Main} & 0.03137 & 1.954e3 & 1.58\\
	& FLM-DNN & 0.01883 & 1.421e3 & 1.60\\
	\hline
\end{tabular}
\label{table:Vanilla RNN}
\end{table}
\begin{table}[!h]
\caption{Performance comparison of algorithms in GRU Network trained with MNIST.
}
\centering
\begin{tabular}{c c c c c}
	\hline
	Network & Method & Memory (Mb) & FLOPs & ER (\%)\\
	\hline
	GRU & Original \cite{GRU}  & 7.11 & 1.77e7 & 0.85 \\
	& SMF \cite{Main} & 0.0327 & 6.21e4 & 1.18\\
	& FLM-DNN & 0.01933 &5.49e4 & 1.20\\
	\hline
\end{tabular}
\label{table:GRU}
\end{table}
To train LSTM networks, the PTB dataset \cite{PTB} is used. Words are fed into the network, which predicts the sequential words in each sentence. "Perplexity" is the evaluation metric we use in this scenario \cite{Perplexity}. In this experiment, two LSTM networks consisting of ten layers including one embedding, four input-to-hidden, four hidden-to-hidden, and one hidden-to-output layer are considered. Two LSTM cells are used in each network. What distinguishes these networks is the number of inputs and representation dimensions, which are 650 and 1500 for each network. $L_1$ norm regularization is adopted and all layers except the hidden-to-output layers are factorized.
\begin{table}[!h]
\caption{Comparison between memory, FLOPs, and perplexity of two LSTM networks trained with different methods.}
\centering
\begin{tabular}{c c c c c}
	\hline
	Network & Method & Memory (Mb) & FLOPs & PPL \\
	\hline
	LSTM-650-650 & Original \cite{LSTM1} & 632.32 & 3.33e7 & 82.7 \\
	& SMF \cite{Main} & 101.33 & 16.0e6 & 92.92\\
	& FLM-DNN &87.98 & 13.79e6 & 93.51\\
	\hline
	LSTM-1500-1500 & Original \cite{LSTM1} & 1568 & 1.38e8&78.29 \\
	& SMF \cite{Main} & 346.26 & 3.95e7 & 88.93\\
	& FLM-DNN & 294.34 & 3.26e7 & 90.12\\
	\hline
\end{tabular}
\label{table:LSTM}
\end{table}
\section{Conclusion}

In this paper, we proposed the FLM-DNN method by implementing binary factorization of weight matrices in deep neural networks. Due to the properties of binary factorization, such as high representational capability and incorporating binary parameters as well as real-valued ones,  FLM-DNN demonstrates an impressive ability to  reduce the computational cost of deep neural networks. Most importantly, we showed that along with this merit,  the network preserves its accuracy.
To evaluate FLM-DNN, ten different networks with different structures were used. The effectiveness of this algorithm was  shown in comparison to some state-of-the-art methods such as Sparse Matrix Factorization, Progressive Pruning, Variational Dropout, and Dynamic Network Surgery. Also, the results suggested that applying binary factorization to weight matrices that belong to more computationally expensive layers   leads to more compression and resource efficiency. This was especially highlighted in ResNet and LSTM networks, where an almost 20\% reduction in the required memory and  a 15\% reduction in the number of FLOPs were achieved.

\newpage

\vfill


\begin{thebibliography}{1}
\bibliographystyle{IEEEtran}

\bibitem{ImageNet}
A. Krizhevsky, I. Sutskever, and G. E. Hilton, ``ImageNet classification with deep convolutional neural networks,''  in \textit{Proc. Adv. Neural Inf. Process. Syst., }, 2012, pp. 1097-1105.


\bibitem{ref2}
K. Simonyan and A.Zisserman, ``Very deep convolutional networks for large-scale image recognition,''  in \textit{Proc. Int. Conf. Learn. Represent., }, 2015, pp. 1-14.

\bibitem{ResNet}
K. He, X. Zhang, S. Ren, and J.Sun, ``Deep residual learning for image recognition,''  in \textit{Proc. IEEE. Conf. Comput. Vis. Pattern Recognit., }, Jun 2016, pp. 770-778.


\bibitem{LSTM1}
S. Hochreiter and J. Schmidhuber, ``Long short-term memory,''  in \textit{Neural Comput., }, vol. 9, no. 8, pp. 1735-1780, 1997.

\bibitem{LSTM2}
A. Graves, A. R. Mohamed, and G. Hinton, ``Speech recognition with deep recurrent neural networks,''  in \textit{Proc. IEEE Int. Conf. Acoust. Speech Signal Process., }, May 2013, pp. 6645-6649.

\bibitem{Pruning1}
S. Han J. Pool, J. Tran, and W. J. Dally, ``Learning both weights and connections for efficient neural network,''  in \textit{Proc. Adv. Neural Inf. Process. Syst. Speech Signal Process., }, 2015, pp. 1135-1143.

\bibitem{Pruning2}
S.Anwar, K. Hwang, and W. Sung, ``Structured pruning of deep convolultional neural networks,''  in \textit{ACM J. Emerg. Technol. Comput. Syst. }, May vol. 13, no. 3, 2017.

\bibitem{Regularization}
S. Scardapane, D. Comminiello, A. Hussain, and A. Uncini, ``Group sparse regularization for deep neural networks,'' in \textit{Neurocomputing}, vol. 241, pp. 81-89, Jun. 2017.

\bibitem{Networksurgery}
Y. Guo, A. Yao, and Y. Chen, ``Dynamic network surgery for efficient DNNs,'' in \textit{Proc. Adv. Neural Inf. Process. Syst.}, 2016, pp. 1379-1387, Jun. 2017.

\bibitem{StructuredSparsity}
W. Wen, C. Wu, Y. Wang, Y. Chen, and H. Li,``Learning structured sparsity in deep neural networks,'' in \textit{Proc. Adv. Neural Inf. Process. Syst.}, 2016, pp. 2074-2082, Jun. 2017.


\bibitem{Fac1}
E. L. Denton, W. Zeremba, J. Bruna, Y. LeCun, and Y. Fergus,``Exploiting linear structure within convolutional networks for efficient evaluation,'' in \textit{Proc. Adv. Neural Inf. Process. Syst.}, 2014, pp. 1269-1277, Jun. 2017.

\bibitem{Fac2}
X. Zhang, J. Zou, X. Ming, K. He, J. Sun,``Efficient and accurate approximations of nonlinear convolutional networks,'' in \textit{Proc. IEEE Conf. Comput. Vis. Pattern Recognit.}, 2015, pp. 1984-1992.

\bibitem{Fac3}
C. Tai, T. Xiao, Y. Zhang, X. Wang, and E. Weinan,``Convolutional neural networks with low-rank regularization,'' in \textit{Proc. Int. Conf. Learn. Represent. Pattern Recognit.}, 2016, pp. 1-11.

\bibitem{Fac4}
M. Jadeberg, A. Vedaldi, A. Zisserman,``Speeding up convolutional neural networks with low rank expansions,'' in \textit{Proc. Brit. Mach. Vis. Conf. Pattern Recognit.}, 2014, pp. 1-12.

\bibitem{Fac5}
K. Ullrich, E. Meeds, and M. Welling,``Soft weight-sharing for neural network compression,'' in \textit{Proc. Int. Conf. Vis. Learn. Represent.}, 2017, pp. 1-16.

\bibitem{Fac6}
D. Molchanov, A. Ashukha, and D. Petrov,``Variational Dropout sparsifies deep neural networks,'' in \textit{34th Int. Conf. Mach. Learn.}, vol. 70, 2017, pp. 2498-2507.

\bibitem{Main}
K. Wu, Y. Guo, C. Zhang,``Compressing deep neural networks with sparse matrix factorization,'' in \textit{IEEE transactions on neural networks and learning systems}, 2019.


\bibitem{Binary1}
R. Kueng, and J. A. Troppe,``Binary component decomposition part I: the positive-semidefinite case,'' in \textit{SIAM Journal of Mathematics of Data Science}, vol. 3, iss. 2, 2021.

\bibitem{Binary2}
R. Kueng, and J. A. Troppe,``Binary component decomposition part II: the asymmetric case,'' in \textit{arxiv}, 2021.

\bibitem{Bi-Real}
Z. Liu, B. Wu, W. Luo, X. Yang, W. Liu, and K. T. Cheng,``Bi-Real Net: Enhancing the Performance of
1-bit CNNs With Improved Representational
Capability and Advanced Training Algorithm,'' in ECCV, 2018.

\bibitem{MNIST}
Y. LeCun, L. Bottou, Y. Bengio, and P. Haffner,``Gradient-based learning applied to document recognition,'' in\textit{Prpc. IEEE}, vol. 86, no. 11, pp. 2278-2324, Nov. 1998.


\bibitem{CIFAR}
A. Krizhevsky and G. Hinton,``Learning multiple layers of features from tiny images,'' in Univ. Toronto, Toronto, ON, Canada, Tech. Rep., 2009, vol. 1, no. 4.

\bibitem{ImageNet2}
O. Russakovsky \textit{et al.},``ImageNet large scale visual recognition challenge,'' in \textit{Int. J. Comput. Vis.}, vol. 115, no. 3, pp. 211-252, 2015.

\bibitem{PTB}
M. P. Marcus, M. A. Marcinikiewicz, and B. Sanorini,``Building a large annoted corpus of English: The Penn treebank,'' in \textit{Comput. Linguistics,}, vol. 19, no. 2, pp. 313-330, 1993.

\bibitem{Caffe}
	Yangoing Jia, Evan Shelhamer, Jeff Donahue, Sergey Karayev, Jonathan Long, Ross Girshick, Sergio Guadarrama, and Trevor Darrell, ``Caffe: Convolutional Architecture for Fast Feature Embedding, '' in \textit{arXiv preprint arXiv:1408.5093}, 2014.
	
\bibitem{VGG-Like}
Yangoing Jia, Evan Shelhamer, Jeff Donahue, Sergey Karayev, Jonathan Long, Ross Girshick, Sergio Guadarrama, and Trevor Darrell, ``Caffe: Convolutional Architecture for Fast Feature Embedding, '' in \textit{arXiv preprint arXiv:1408.5093}, 2014.


\bibitem{Adam}
D. P. Kingma and J. Ba,`Adam: A method for stochastic optimization,'' in \textit{Proc. Int. Conf. Learn. Represent.,} 2015, pp. 1-15.


\bibitem{VGG}
Liu, Shuying and Deng, Weihong,`Very deep convolutional neural network based image classification using small training sample size,'' in \textit{2015 3rd IAPR Asian Conference on Pattern Recognition (ACPR),} 2015, pp. 730-734.


\bibitem{Perplexity}
D. Newman, E. V. Bonilla, and W. Buntine,`Improving topic coherence with regularized topic method,'' in \textit{Adv. Neural Inf. Process. Syst.,} 2011, pp. 496-504.

\bibitem{Binarized}
Matthieu Courbariaux, Itay Hubara, Daniel Soudry, Ran El-Yaniv, and Yoshua Bengio,`Binarized Neural Networks: Training Neural Networks with Weights and
Activations Constrained to +1 or \-1,'' in \textit{arXiv:1602.02830v3,} 2016.

\bibitem{Real-to-Binary-Convolutions}
Brais Martinez, Jing Yang, Adrian Bulat, and Georgios Tzimiropoulos ,`TRAINING BINARY NEURAL NETWORKS WITH REAL-TO-BINARY CONVOLUTIONS,'' in \textit{ICLR,} 2020.


\bibitem{XORNet}
M. Rastegari, V. Ordonez, J. Redmon, A. Farhadi ,`Xnor-net: Imagenet classification using binary convolutional neural networks,'' in \textit{ECCV,} 2016.



\bibitem{LQNet}
D. Zhang, J. Yang, D. Ye, G. Hua, Lq-nets ,`Learned quantization for highly accurate and compact deep
neural networks,'' in \textit{ECCV,} 2018.


\bibitem{Distilled}
X. Chen, G. Liu, J. Shi, J. Xu, and B. Xu ,`Distilled binary neural network for monaural speech separation,'' in \textit{IJCNN,} 2018.

\bibitem{Distillation and Quantization}
A. Polino, R. Pascanu, and D. Alistarh ,` Model compression via distillation and quantization,'' in \textit{ICLR,} 2018.


\bibitem{Apprentice}
A. Mishra and D. Marr ,` Apprentice: Using knowledge distillation techniques to improve low-precision network
accuracy,'' in \textit{ICLR,} 2018.

\bibitem{Circulant}
C. Liu, W. Ding, X. Xia, B. Zhang, J. Gu, J. Liu, R. Ji, and D. Doermann,` Circulant binary convolutional
networks: Enhancing the performance of 1-bit dcnns with circulant back propagation,'' in \textit{IEEE CVPR,} 2019.



\bibitem{Half-wave}
Z. Cai, X. He, J. Sun, and N. Vasconcelos,` Deep learning with low precision by half-wave gaussian quantization,'' in \textit{IEEE CVPR,} 2017.


\bibitem{Mixed-Precision}
Sharan Narang, Gregory Diamos, Erich Elsen, Paulius Micikevicius, Jonah Alben, David Garcia,
Boris Ginsburg, Michael Houston, Ganesh Venkatesh, and Hao Wu,` Mixed precision training,'' in \textit{arXiv preprint arXiv:1710.03740,} 2017.

\bibitem{NVIDIA}
Luke Durant, Olivier Giroux, Mark Harris, and Nick Stam,` NVIDIA Tesla V100 GPU architecture,'' 2017. URL https://devblogs.nvidia.com/inside-volta/.

\bibitem{ConvFac}
C. Tai, T. Xiao, Y. Zhang, X Wang, and E. Weinan,` Convolutional neural networks with low-rank regularization,'' in \textit{Proc. Int. Conf.Learn. Represent.,} 2016, pp. 1-11.

\bibitem{RNN}
Rumelhart, David E; Hinton, Geoffrey E, and Williams, Ronald J,` Learning internal representations by error propagation,'' in \textit{Tech. rep. ICS 8504.} San Diego, California: Institute for Cognitive Science, University of California, 1985, pp. 1-11.

\bibitem{GRU}
Kyunghyun Cho, Bart van Merrienboer, Dzmitry Bahdanau, and Yoshua Bengio,` On the Properties of Neural Machine Translation: Encoder-Decoder Approaches,'' in \textit{	arXiv:1409.1259}, 2014.

\bibitem{VGG-Like Github}
Kyunghyun Cho, Bart van Merrienboer, Dzmitry Bahdanau, and Yoshua Bengio,` On the Properties of Neural Machine Translation: Encoder-Decoder Approaches,'' in \textit{	arXiv:1409.1259}, 2014.

\end{thebibliography}
\end{document}